 \documentclass[sigconf]{acmart}

\AtBeginDocument{%
  }

\acmConference[STIntelligence, CIKM '25]{The International Workshop on Spatio-Temporal Data Intelligence and Foundation Models, The 34th ACM International Conference on Information and Knowledge Management}{November 10--14, 2025}{Seoul, Republic of Korea}

\begin{document}

\title{NextLocLLM: Location Semantics Modeling and Coordinate-Based Next Location Prediction with LLMs}


\author{Shuai Liu}
\email{shuai.liu@ntu.edu.sg}
\affiliation{%
  \institution{Nanyang Technological University}
  \city{Singapore}
  \country{Singapore}
}

\author{Ning Cao}
\email{ning.cao@ntu.edu.sg}
\affiliation{%
  \institution{Nanyang Technological University}
  \city{Singapore}
  \country{Singapore}
}

\author{Yile Chen}
\email{yile001@e.ntu.edu.sg}
\affiliation{%
  \institution{Nanyang Technological University}
  \city{Singapore}
  \country{Singapore}
}

\author{Yue Jiang}
\email{yue013@e.ntu.edu.sg}
\affiliation{%
  \institution{Nanyang Technological University}
  \city{Singapore}
  \country{Singapore}
}

\author{George Rosario Jagadeesh}
\email{george.jagadeesh@stengg.com}
\affiliation{%
  \institution{ST Engineering}
  \city{Singapore}
  \country{Singapore}
}

\author{Gao Cong}
\email{gaocong@ntu.edu.sg}
\affiliation{%
  \institution{Nanyang Technological University}
  \city{Singapore}
  \country{Singapore}
}

\renewcommand{\shortauthors}{Liu.S et al.}


\begin{abstract}
  Next location prediction is a critical task in human mobility analysis. 
  Existing methods typically formulate it as a classification task based on discrete location IDs, which hinders spatial continuity modeling and limits generalization to new cities.
In this paper, we propose NextLocLLM, a novel framework that reformulates next-location prediction as coordinate regression and integrates LLMs for both location semantics encoding and coordinate-level prediction.
To model location functional semantics, it constructs LLM-enhanced POI embeddings by leveraging  language understanding capabilities of LLMs to extract functional semantics from textual descriptions of POI categories.
These POI embeddings are combined with spatiotemporal trajectory representation and fed into the same LLM, enabling unified semantic and predictive modeling.
A lightweight regression head generates coordinate outputs, which are mapped to top-k candidate locations via post-prediction retrieval module, ensuring structured outputs.
Experiments across diverse cities show that NextLocLLM outperforms existing baselines in both supervised and zero-shot settings.
Code is available at: \url{https://github.com/liuwj2000/NexelocLLM}.

\end{abstract}
\keywords{Next location prediction,  Large language models (LLMs), Zero-shot learning, Cross-city generalization, LLM-enhanced POI embeddings}

\maketitle

\section{Introduction}
The widespread adoption of location-aware devices and positioning technologies has led to an abundance of human mobility data.
This data supports a wide range of downstream applications, including traffic forecasting~\cite{medina2022urban,kraemer2020effect}, epidemic control~\cite{ceder2021urban}, and location-based personalized services~\cite{lian2020geography}.
At the core of these applications lies the  next-location prediction, which aims to forecast a user's likely future destination based on his/her past trajectories~\cite{chekol2022survey, rajule2023mobility,zhang2018mobility}.

Existing approaches for next-location prediction predominantly rely on deep learning techniques,  with
RNN- and Transformer-based models~\cite{li2024mcn4rec,yao2017serm,kong2018hst,sun2024going,rao2024next} widely used to capture sequential dependencies.
More recently, large language models (LLMs)\cite{achiam2023gpt,touvron2023llama}  have been explored for mobility prediction \cite{wang2023would,liang2024exploring,beneduce2024large,feng2024agentmove}.
Despite  their architectural diversity, most of these methods formulate next location prediction as a classification task over discrete location IDs, which leads to several key limitations. 
First,  location IDs are city-specific and inconsistent   across cities, making it difficult for models trained in one city to generalize to another~\cite{jiang2021transfer}.
Second,  the set of candidate locations is fixed during training, preventing the model from handling unseen locations at inference~\cite{jiang2021transfer}.
Third, dividing space into discrete location IDs  breaks spatial continuity and hinders the ability to capture geographic proximity and relative positions~\cite{liu2016predicting}.
Furthermore, deep learning methods typically rely only on structured, numerical features, ignoring  rich and informative textual information such as POI category descriptions.
This limits their ability to infer user intentions from semantic context and to generalize across semantically similar locations.
Prompt-based LLM models attempt to address these issues but face two additional challenges: (1) their outputs may violate structural constraints, producing more or fewer than the required number of predictions, even when explicitly specified in the prompt; and (2) their sequential, autoregressive decoding mechanism leads to slow inference~\cite{wang2023would,liang2024exploring}.
These limitations highlight the need for a new method that  better models spatial relationships, generalizes effectively across cities, and leverages semantic information, while maintaining structured outputs and inference efficiency.

\begin{figure*}[h]
\centering
\includegraphics[scale=0.4]{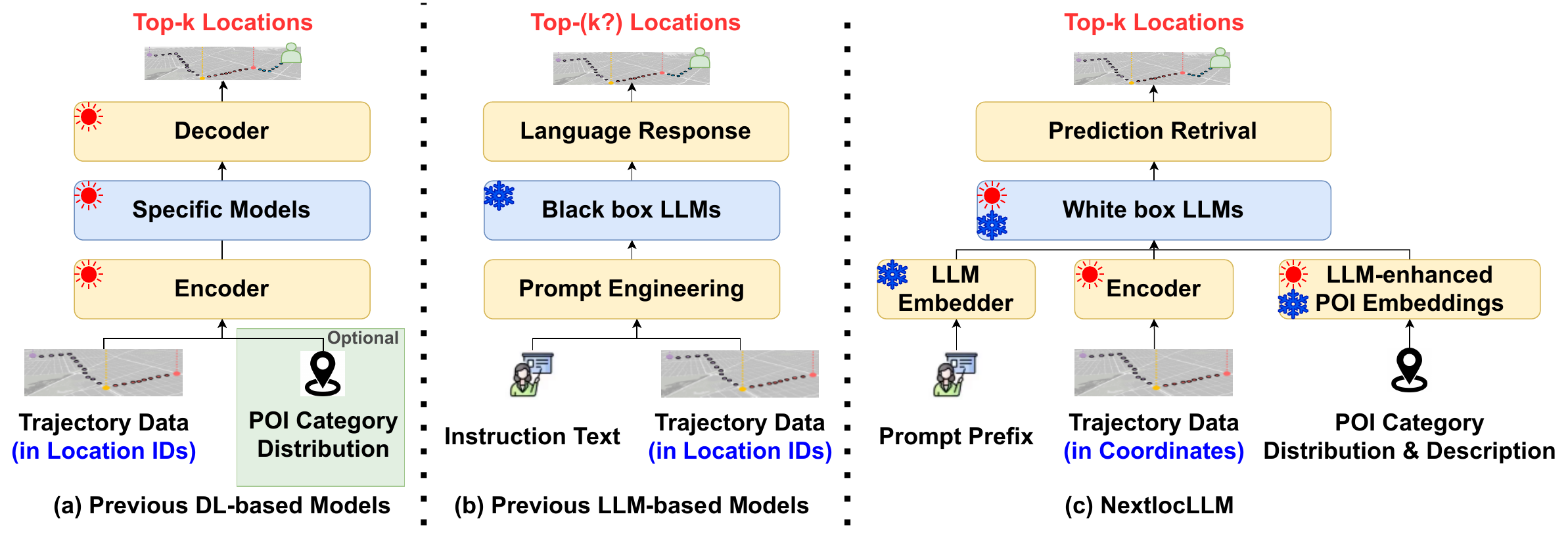}

\caption{Comparison between deep-learning (DL)-based models, previous LLM-based models, and NextlocLLM.  Black-box LLMs are typically accessed through APIs. White-box LLMs mean models where their parameters are directly accessible.}
\label{fig:model vs}
\end{figure*}

To address the above limitations, we propose NextLocLLM, a novel framework that reformulates next-location prediction as coordinate regression and structurally integrates large language model (LLM) for both the location-semantic encoder and the prediction backbone.
Instead of  classifying over discrete location IDs, NextLocLLM directly predicts normalized geographic coordinates, which naturally preserves spatial continuity and allows for predicting previously unseen locations.
To incorporate semantic context and capture the functional characteristics of each location, NextLocLLM constructs LLM-enhanced POI embeddings  by leveraging the natural language understanding capabilities of LLMs to encode textual descriptions of POI categories, which are then aggregated according to location-specific distributions.
These semantic embeddings are combined with spatial temporal trajectory embeddings and passed through a partially frozen LLM backbone, which preserves pre-trained knowledge while adapting to the prediction task.
A lightweight regression head maps LLM's final hidden representation to coordinates, representing the next predicted location.
By using the same LLM backbone for both POI semantics embedding and next-location prediction, NextLocLLM unifies POI descriptions and trajectory context within a consistent representation space.
This enables LLM to align semantic meaning across cities and to interpret user intent based on both spatial temporal patterns and functional semantics.
During inference, a post-prediction retrieval module maps predicted coordinates to top-k candidate locations, ensuring structured outputs for evaluation and deployment.
As illustrated in Figure~\ref{fig:model vs}, NextLocLLM differs from prior deep learning and prompt-based LLM models. Extensive experiments demonstrate that it outperforms existing baselines under both fully supervised and zero-shot settings.
In conclusion, our contributions are:

\begin{itemize}
    \item We introduce NextLocLLM, a novel framework that reformulates next-location prediction as coordinate regression and integrates LLMs for both semantic encoding and coordinate-level prediction.
    This formulation preserves spatial continuity and enables generalization  across cities.
   
    \item We introduce LLM-enhanced POI embeddings, which leverage the natural language understanding capabilities of LLMs to capture functional semantics from textual descriptions of POI categories.

    \item Extensive experiments on multiple real-world datasets demonstrate the strong performance of NextLocLLM under both fully supervised and zero-shot settings.
\end{itemize}

\section{Related Work}
Next location prediction serves as a core problem in mobility modeling.
Early methods  relied on handcrafted features and statistical models. 
They applied domain-specific signals like temporal patterns, geographic distances, and location transitions. 
\citep{noulas2012mining}
proposed features capturing transitions between locations  and spatiotemporal check-in patterns. 
\citep{chen2014nlpmm}  introduced a Markov-based model combining individual and collective mobility behaviors. 
\citep{ying2014mining} utilized geographic, temporal, and semantic signals. 
While being foundational, they suffered from limited generalizability and scalability due to their reliance on manual feature design.

With the rise of deep learning,  RNNs and Transformers have become mainstream. 
~\cite{wang2023towards}  incorporates event embeddings to represent routine behaviors and disruptions.
\citep{fan2018deep} combined CNNs and bidirectional LSTMs to integrate contextual information for prediction.
TrajGraph~\cite{zhao2024trajgraph} employs a graph Transformer to efficiently encode spatiotemporal context under reduced computational complexity.
GETNext~\citep{yang2022getnext} introduced a Graph Enhanced Transformer that utilized global trajectory flow map.
AGCL~\cite{rao2024next} introduces a multi-graph learning framework with adaptive POI graphs, spatial-temporal attention, and bias correction. 
iPCM~\cite{song2025integrating} combines global trajectory with personalized user embeddings using a Transformer encoder and probabilistic correction module.
MCN4Rec~\cite{li2024mcn4rec} takes a multi-perspective approach, collaboratively learning from both local and global views to model heterogeneous relationships among users, POIs, temporal factors, and activity types.
STMGCL~\cite{jia2023improving} introduces temporal group contrastive learning within a self-attention encoder to uncover user preference groups.
MCLP~\cite{sun2024going} leverages topic models to extract latent user preferences and enhances arrival time estimation via attention.
CLLP~\cite{zhou2024cllp} fuses local and global spatiotemporal contexts to track evolving user interests.
FHCRec~\cite{chen2025enhancing} captures both long- and short-term patterns through hierarchical contrastive learning over subsequences.
SanMove~\cite{wang2023sanmove} proposes a non-invasive self-attention module that utilizes auxiliary trajectory signals to learn short-term preferences. 
LoTNext~\cite{xu2024taming} addresses the long-tail challenge via graph and loss adjustments that rebalance POI interaction distributions.

Recent advancements in LLMs~\citep{touvron2023llama,touvron2023llama1,achiam2023gpt} have opened new possibilities for next location prediction. 
Current studies typically use prompt engineering to reformulate trajectories and prediction tasks as natural language inputs.
\citep{wang2023would} introduced the concepts of historical and contextual stays to capture long- and short-term dependencies in human mobility,  incorporating time-awareness into predictions. 
\citep{wang2024large} leveraged the semantic perception capabilities of LLMs to extract personalized activity patterns from historical data.
AgentMove~\cite{feng2025agentmove} decomposes the next-location prediction task into three specialized components: a spatial-temporal memory module that captures individual behavioral patterns, a world knowledge generator that infers structural and urban influences, and a collective knowledge extractor that models shared mobility patterns. 
CausalMob~\cite{yang2024causalmob} introduces a causality-inspired framework that leverages LLMs to extract latent intention signals tied to external events.
Despite these advances, most models  frame next-location prediction as a classification task over discrete location IDs, which limits their prediction accuracy and generalization ability.

\section{PROBLEM FORMULATION}
Let $\mathcal{L}=\{loc_1,\cdots,loc_p\}$ be the set of locations, where each location is represented as a tuple $(\text{id},x,y,\text{poi})$: a unique identifier $\text{id}$, spatial coordinates of the location's centroid $(x,y)$, and corresponding POI attributes $\text{poi}$.  
The temporal context is given by $\mathcal{TI}=\{(d,t)\}$, where $d$ is day-of-week and $t$ is hour-of-day.

\textit{Definition 2.1 (\textbf{Visiting Record}).} A visiting record is defined as a tuple $s=(loc,(d,t),dur)$, indicating that a user visits location $loc$ on day $d$ at hour $t$, and stayed for a duration of $dur$.

\textit{Definition 2.2 (\textbf{POI Attributes}).}We treat each location as a region composed of multiple POIs.
POI attributes for each location $loc$ are represented as $poi=(intr,freq)$, where $intr=(i_1,\cdots,i_r)$ denotes the natural language descriptions of the $r$ POI categories, and $freq=(f_1,\cdots,f_r)$ represents the MinMax-normalized values of each POI category for that location, reflecting the relative strength of each category.
Here $r$ is the number of POI categories.

\textit{Definition 2.3 (\textbf{Historical and Current Trajectories}).} A user's mobility trajectory is divided into a historical trajectory $S_h=(s_{t-M+1},\cdots,s_{t})$ and a current trajectory $S_c=(s_{t+1},\cdots,s_{t+N})$.
Unless otherwise specified, we will use $l_{seq}$ to refer to either $N$ or $M$ in certain contexts of subsequent sections.

\textit{Definition 2.4 (\textbf{Next Location Prediction}).} 
Given a user's historical trajectory $S_h$ and current trajectory $S_c$, the target is to predict the {ID of} next location $loc_{t+1}$  that the user is most likely to visit.

\begin{figure*}[h]
\centering
\includegraphics[scale=0.3]{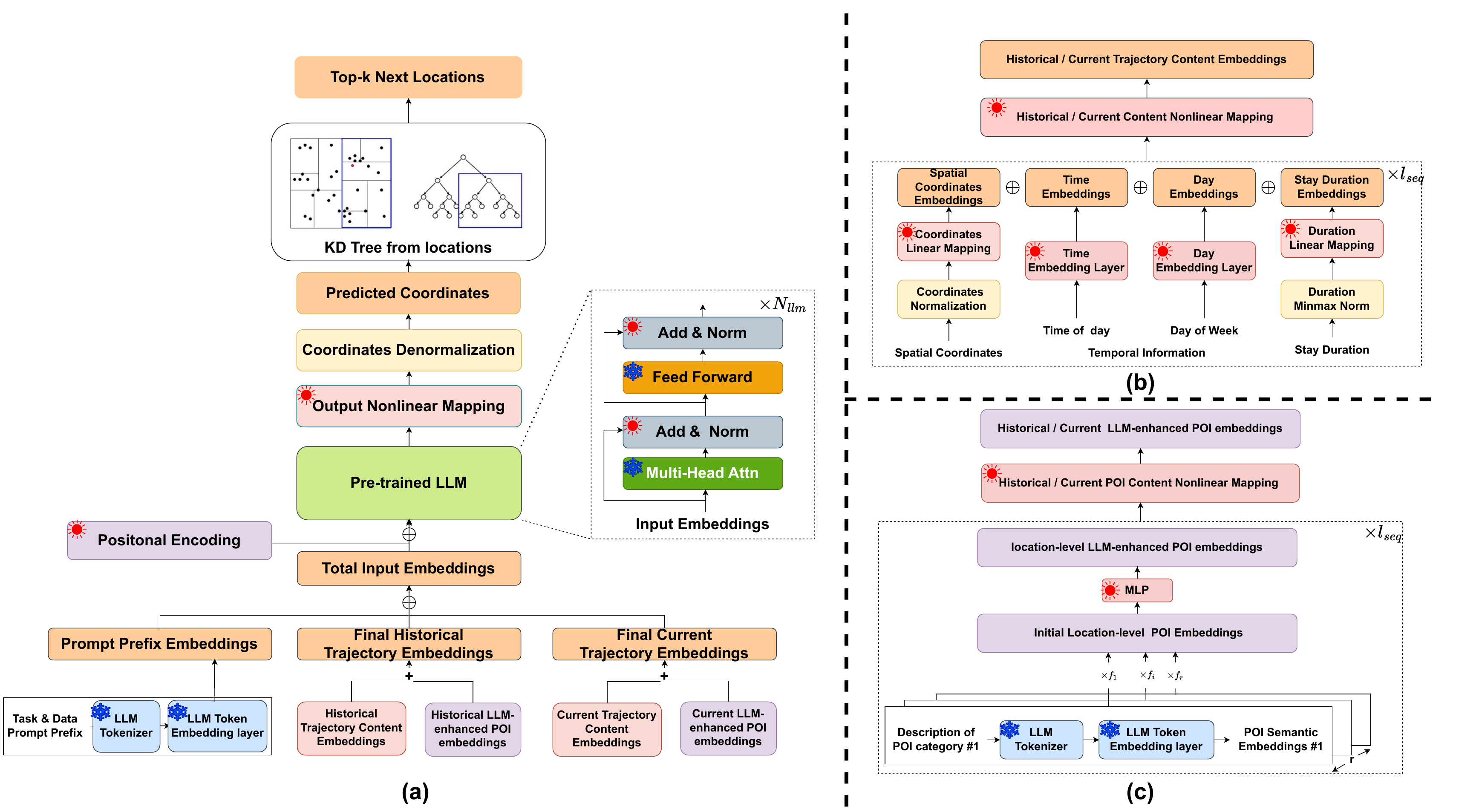}

\caption{Structure of NextlocLLM. (a) shows its overall structure, (b) represents spatial temporal trajectory embeddings, and (c) demonstrates the LLM-enhanced POI embeddings.}
\label{fig:model}
\end{figure*}

\section{Method}
Figure~\ref{fig:model} illustrates an overview of  NextLocLLM, a regression-based next-location prediction framework that leverages a unified LLM backbone for both location semantic representation and coordinate-level next location prediction.
It consists of four main components: (1) spatial-temporal trajectory embeddings, (2) LLM-enhanced POI embeddings, (3) LLM-based coordinate predictor, and (4) a post prediction retrieval module.
It begins by embedding spatial coordinates, temporal features (day and hour), and stay durations independently. 
These embeddings are then concatenated to form the input representations for historical and current trajectories.
To capture the functional characteristics of locations, we construct LLM-enhanced POI embeddings by encoding the textual descriptions of POI categories, using LLMs' natural language understanding capabilities. 
These semantic vectors are aggregated using a weighted average, where the weights are derived from normalized POI category counts, indicating the relative strength of each function type.
Next, the trajectory embeddings, POI embeddings, and a task-specific prompt prefix are combined and fed into the same LLM backbone. 
The final hidden representation from the LLM is passed through a lightweight regression head, which maps it to a pair of coordinates representing the predicted next location.
During inference, these predicted coordinates are mapped to top-k candidate locations through a post-prediction retrieval module, resulting in structured outputs suitable for evaluation and deployment.

\subsection{Spatial-temporal trajectory embeddings}
We extract spatial-temporal features from  historical trajectory $S_h$ and  current trajectory $S_c$, including spatial coordinates $(x,y)$, time-of-day $T$, day-of-week $D$ and stay duration $Dur$.
For each feature type, the same embedding function is applied to $S_h$ and $S_c$.

\noindent \textbf{Temporal Embeddings}.
To capture temporal characteristics, we employ independent embedding tables for  time-of-day ($T$) and day-of-week ($D$).
These temporal signals are essential, as user mobility patterns often vary significantly across different hours and days.
We embed each temporal feature via lookup operations on learnable embedding tables, yielding time-of-day embeddings $E_T=f_t(T) \in \mathbb{R}^{l_{seq} \times d_t}$ and day-of-week embeddings $E_D=f_d(D)\in \mathbb{R}^{l_{seq} \times d_d}$.

\noindent \textbf{Stay Duration Embeddings}
Stay duration represents another important temporal factor that reflects user intent. 
Longer stays typically correspond to complex activities (e.g., working, dining), while shorter stays are associated with quick or transitional actions.
We first normalize the raw duration values using min-max scaling to obtain $Dur'$, and then project them via a linear transformation: $E_{Dur}=f_{dur}(Dur')\in \mathbb{R}^{l_{seq} \times d_{dur}}$.

\noindent \textbf{Spatial Coordinates Embeddings}.
Existing methods typically rely on discrete location IDs  to represent spatial information.
However, these discrete IDs fail to capture the geographical relationships between locations and hinder generalization across cities.
To address this, we represent locations using spatial coordinates of their centroids instead of discrete IDs. 
Specifically, we utilize Web Mercator coordinates, as they can reflect the spatial relationships between locations. 
At  city level, the gap between geodesic distance and Euclidean distance under Web Mercator projection is minimal, making it suitable for practical urban mobility tasks~\citep{battersby2014implications,peterson2014mapping}.
While spatial coordinates effectively capture spatial relationships within cities, their raw ranges vary significantly between cities, posing a challenge for generalization in zero-shot scenarios.
To mitigate this, we normalize the coordinates by scaling them to a standard normal distribution $N(0,1)$.
The mean and variance for it are based on spatial coordinates from all trajectory visiting records , rather than from its geometric map,
as areas with high population density often do not align with the city's geometric center~\cite{yabe2024yjmob100k}.
The normalized coordinates $(x',y')$ are then transformed into spatial coordinates embeddings via a linear function: 
\begin{equation}
    E_{XY}=f_{xy}(x',y')\in \mathbb{R}^{l_{seq} \times d_{xy}}.
\end{equation}
Another aspect to note is that NextLocLLM  treats mobility individuals as free-moving particles within the urban space, unconstrained by road networks. 
This design choice enhances the model’s spatial flexibility and improves its transferability across different cities.

\noindent \textbf{Trajectory Content Embeddings}.
\label{sec: trajectory content embedding}
So far, we have obtained four types of embeddings: spatial coordinates embeddings $E_{XY}$, time-of-day embeddings $E_{T}$, day-of-week embeddings $E_{D}$, and stay duration embeddings $E_{Dur}$. 
To integrate these features,  we concatenate the embeddings for each timestep within a trajectory to form the combined content embeddings:
\begin{align}
    E_{all_h}&=E_{XY_h}||E_{T_h}||E_{D_h}||E_{Dur_h} \in \mathbb{R}^{M\times (d_{xy}+d_{t}+d_d+d_{dur})},\\
    E_{all_c}&=E_{XY_c}||E_{T_c}||E_{D_c}||E_{Dur_c} \in \mathbb{R}^{N\times (d_{xy}+d_{t}+d_d+d_{dur})}.
\label{eq:1}
\end{align}
After we obtain $E_{all_h}$ and $E_{all_c}$, we apply nonlinear transformations $f_h$ and $f_c$, implemented as multi-layer perceptrons (MLPs), to project these embeddings into the LLM’s embedding dimension $d_{llm}$:
\begin{equation}
    E_{con_h}=f_h(E_{all_h}) \in \mathbb{R}^{M\times d_{llm}},
    E_{con_c}=f_c(E_{all_c}) \in \mathbb{R}^{N\times d_{llm}}.
\end{equation}

\subsection{LLM-enhanced POI Embedding}
Understanding the functional attributes of a location is essential for accurate next-location prediction, as user mobility behaviors are often correlated with the underlying semantics of their visited locations.
However, traditional deep learning models typically rely on numerical inputs like location IDs, spatial coordinates, and timestamps, without incorporating any information about the types or functions of locations.
While some prior works have attempted to integrate POI information~\cite{al2025seaget,yang2022getnext} to model location functions, they generally do so using discrete category IDs or frequency vectors, which fail to capture the rich semantics embedded in POI textual descriptions.
In practice, POIs often come with rich natural-language metadata, such as category names or short descriptions, that provide high-level cues about a location’s functionality.
These textual descriptions carry transferable knowledge across cities, unlike raw POI frequency distribution vectors, which are often specific to local POI taxonomies and lack semantic alignment.

To fully exploit the semantic signals within POI metadata and better model the functional attributes of locations,  we propose LLM-enhanced POI embeddings, which leverage the language understanding capabilities of LLMs.
Given the natural-language descriptions of POI categories $intr=(i_1,\cdots,i_r)$, we obtain semantic embeddings $E_I=(E_1,\cdots,E_r)\in \mathbb{R}^{r\times l\times d_{llm}}$  using the token embedding layer of the LLM, where 
$l$ denotes the maximum tokenized description length and $d_{llm}$  is the embedding dimension.
To aggregate these semantic embeddings into a location-level embedding, we use normalized POI category counts $freq=(f_1,\cdots,f_r)$, which reflects each category’s relative strength across locations, as weights and compute a weighted sum of the semantic vectors:
\begin{equation}
    E_{POI_{loc,init}}=\sum_{j=1}^r E_j * f_j \in \mathbb{R}^{lt \times d_{llm}}.
\end{equation}
We then apply a nonlinear transformation via a multi-layer perceptron to obtain the location-level LLM-enhanced POI embedding: 
\begin{equation}
    E_{POI_{loc}}=\mathrm{MLP}(E_{POI_{loc,init}}) \in \mathbb{R}^{d_{llm}}.
\end{equation}
For each visiting record, we attach the corresponding location-level POI embedding, producing the initial trajectory-level POI embedding sequence $E_{L-POI} \in \mathbb{R}^{l_{seq} \times d_{llm}}$.
To accommodate the distinct temporal roles of historical and current trajectories, we further apply separate MLPs to transform this shared representation into two trajectory-specific forms: 
\begin{align}
    E_{L-POI_h}&=f_{h_{poi}}(E_{L-POI}) \in \mathbb{R}^{M \times d_{llm}},\\ E_{L-POI_c}&=f_{c_{poi}}(E_{L-POI}) \in \mathbb{R}^{N \times d_{llm}}.
\end{align}

\subsection{Large Language Model backbone}
\noindent \textbf{Total Input Embeddings}
Given the the content embeddings $E_{con_h}$ and $E_{con_c}$, along with the  LLM-enhanced POI embeddings $E_{L-POI_h}$ and $E_{L-POI_c}$, we combine each pair through element-wise addition to obtain the final historical and current trajectory embeddings:
\begin{equation}
    E_{his}=E_{con_h}+E_{L-POI_h}, E_{cur}=E_{con_c}+E_{L-POI_c}.
\end{equation}
To improve the model's comprehension of input data and prediction task, we design a task-specific prompt prefix that encodes high-level instructions, describes the input format, and clarifies the role of historical and current trajectories. 
This prefix is tokenized and passed through the LLM’s embedding layer to obtain the prompt embedding $E_{prompt}$, which is concatenated with the trajectory embeddings to form the final input:
\begin{equation}
    E_{total}=E_{prompt}||E_{his}||E_{cur}
\end{equation}

\noindent \textbf{Partially Frozen Large Model}.
To retain the knowledge encoded during pretraining while enabling task adaptation, we adopt a partially frozen strategy inspired by~\citep{zhou2023one}.
Specifically, we freeze the LLM’s self-attention and feedforward layers to preserve semantic priors, while fine-tuning a small set of parameters, such as those in positional encoding and layer normalization~\citep{zhou2023one,liu2024spatial}.
This lightweight tuning approach offers a favorable balance between performance and computational efficiency.

\noindent \textbf{Regression-based Method}
We adopt a regression-based formulation to predict the spatial coordinates of next location, rather than selecting from a predefined set of discrete location IDs.
This formulation removes the dependency on city-specific location ID systems and enables generalization to unseen locations.
When paired with the expressive capacity of LLM, it further allows NextLocLLM to integrate spatial cues with semantic context for more accurate and generalizable predictions.
 Given the total input embedding $E_{total}$, LLM  processes it and produces the output representation $E_o=\mathrm{LLM}(E_{total})$. 
We extract the last vector $\textbf{v}_{o} \in \mathbb{R}^{d_{llm}}$ from $E_o$  and apply an nonlinear mapping  $f_o$ to generate the normalized predicted spatial coordinates $xy_o'=f_o(o)$.
The output is then denormalized to obtain the final spatial prediction $xy_o$ in the original coordinate scale.
During training, the objective is to minimize the Euclidean distance between $xy_o$ and ground truth $\hat{xy}$.

\subsection{Post-Prediction Retrieval Module}
During inference, to generate structured next-location predictions, we incorporate a post-prediction retrieval module.
Specifically, we construct a KD-tree using the spatial coordinates of all candidate locations' centroids.
The predicted coordinates $xy_o$
  serve as the query point for this KD-tree. 
  The retrieval process identifies the top-
$k$ locations nearest to  $xy_o$ in Euclidean space.
These retrieved locations constitute the final top-$k$ predictions, converting the continuous regression output into a discrete format suitable for evaluation and deployment.
Notably, the KD-tree is used only during inference as a lightweight spatial indexing structure to convert continuous coordinate predictions into top-$k$ discrete location candidates.
It does not participate in training and does not affect the prediction accuracy itself. 
Alternative spatial indexing structures such as ball tree  yield nearly identical results, as the accuracy is primarily determined by quality of the predicted coordinates, not the indexing mechanism.

\begin{table*}
  \caption{Fully Supervised Next Location Prediction Result}
  \label{tab:supervise 1}
  \centering
  
  \begin{tabular}{|l|ccc|ccc|ccc|}
    \toprule
    {Method } &  \multicolumn{3}{c|}{Xi'an}   &  \multicolumn{3}{c|}{Singapore}  &  \multicolumn{3}{c|}{Japan}  \\
    & Hit@1 & Hit@5 & Hit@10  & Hit@1 & Hit@5 & Hit@10  & Hit@1 & Hit@5 & Hit@10 \\ 
    \cmidrule(r){1-1}\cmidrule(r){2-4}\cmidrule(r){5-7}\cmidrule(r){8-10}
    STRNN & 11.01\% & 19.15\% & 25.61\% & 3.639\% & 10.23\% & 12.76\%& 8.616\%&15.45\%& 24.46\% \\
    LSTM & 9.753\% & 31.17\% & 45.34\% & 3.197\% & 8.698\%& 10.46\% &2.817\% &9.993\% & 15.67\%  \\
    FPMC & 20.97\% & 39.95\% & 47.58\% & 4.003\% & 11.19\% & 17.63\%& 2.973\%& 7.859\% & 13.51\% \\
    GRU & 9.590\% & 30.92\% & 45.17\% & 2.682\% & 6.051\% & 7.784\%& 3.831\% & 10.69\%& 15.56\% \\
    C-MHSA & 50.32\% & 92.43\% & 95.38\% & 4.874\%& 13.54\% & 19.38\% & \underline{20.17\%}& 30.23\% & 37.68\% \\
    DeepMove & 41.19\% & 83.02\% & 90.85\% & 6.650\% & 20.00\% & 31.08\%& 11.71\%& 22.23\% & 33.35\% \\
    GETNext & 48.63\% & 85.67\% & 93.25\% & 6.498\% & 25.80\% & 32.04\%& 19.17\% & 28.62\% & 33.79\% \\
    SEAGET & 48.79\% & 85.77\% & 93.21\% & 6.512\% & 25.94\% & 32.56\% & 19.26\% & 28.71\% & 33.52\% \\
    CLLP & 47.28\% & 88.37\% & 94.14\% & 6.712\% & 26.98\% & 34.99\% & 17.61\% & 27.76\% & 37.79\%\\
    ROTAN & \underline{52.36\%} & \underline{93.77\%} & \underline{96.25\%} & {6.892\%} & \underline{27.71\%} & \underline{35.56\%} & \textbf{20.23\%} & \textbf{32.76\%} & \underline{43.97\%}\\
    LLMMob & 33.52\% & 77.86\% & 78.00\% & {6.933}\% & 21.07\% & 30.70\% & 17.63\% & 28.55\% & 37.26\% \\
    LLMMob(s) & 20.81\% & 62.08\% & 62.23\% &5.341\% & 17.60\% & 24.47\%& 12.26\%& 21.87\%& 31.19\%\\
    Mobility-LLM & 48.92\% & 88.63\% & 94.26\% & \underline{7.174}\% & 24.43\% & 32.69\% & 14.22\% & 26.71\% & 35.04\% \\
    ZS-NL & 20.92\% & 53.29\% & 66.99\% & 4.199\% & 14.68\% & 20.11\% & 13.07\% & 22.31\% & 26.15\%\\
    ZS-NL(s) & 20.27\% & 52.22\% & 64.97\% & 2.467\% & 5.768\% & 7.104\%& 11.32\% & 19.15\% & 23.57\%  \\
    NextlocLLM & \textbf{58.14\%} & \textbf{97.14\%} & \textbf{99.36\%} & \textbf{7.823}\% & \textbf{30.64\%} & \textbf{36.15\%} & 19.36\% & \underline{31.82\%} & \textbf{46.06\%} \\
    \bottomrule
  \end{tabular}
  
\end{table*}

\section{Experiment}
We evaluate NextLocLLM in both fully-supervised and zero-shot next-location prediction settings. 
In the supervised case, models are trained and tested on the same cities; in the zero-shot setting, evaluation is conducted on unseen cities without retraining.

\subsection{Experimental Setup}
\textbf{Baseline Models.}
We compare NextLocLLM with various recent baseline models, including DL-based models (LSTM~\citep{graves2012long}, FPMC~\citep{rendle2010factorizing}, GRU~\citep{chung2014empirical}, STRNN~\citep{liu2016predicting}, C-MHSA~\citep{hong2023context}, DeepMove~\citep{feng2018deepmove}, GETNext~\citep{yang2022getnext}, SEAGET~\cite{al2025seaget}, CLLP~\cite{zhou2024cllp}, and ROTAN~\cite{feng2024rotan}), and LLM-based models  (LLMMob~\citep{wang2023would}, Mobility-LLM~\cite{gong2024mobility} and ZS-NL~\citep{beneduce2024large}).

\noindent \textbf{Datasets.} We use three datasets: Xi’an~\citep{zhu2023synmob}\footnote{\url{https://github.com/Applied-Machine-Learning-Lab/SynMob}}, Japan~\citep{yabe2024yjmob100k}\footnote{\url{https://zenodo.org/records/13237029}} and  private dataset Singapore. 
Following ~\cite{yabe2024yjmob100k}, we partition each city into 500m × 500m square grids and treat each grid  as a discrete location. 

\noindent \textbf{Evaluation Metrics}.
Following ~\cite{luo2021stan}, we adopt Hit@1/5/10 as evaluation metrics, which measure whether the ground-truth next location falls within the top-1, top-5, or top-10 predicted candidates.

\noindent \textbf{Experimental Configuration}
Each dataset is split into training, validation, and testing with a 70\%/10\%/20\% ratio. 
For zero-shot evaluation, only test set from target city is used, and the post-prediction KD-tree is constructed using the centroid coordinates of candidate locations in the target city.
NextLocLLM uses GPT-2 as its LLM backbone.
Each input sequence includes 40 historical and 5 current records.
All models are trained on Tesla V100 GPUs. 
Due to space constraints, other hyperparameter settings, prompt prefix details, and natural-language POI category descriptions are available in our code repository.

\subsection{Fully Supervised Next Location Prediction Performance}
Table~\ref{tab:supervise 1}  presents the fully-supervised results on Xi’an, Singapore, and Japan datasets. 
RNN-based models (STRNN, LSTM, GRU)  struggle with long-range dependencies, while attention-based models (C-MHSA, DeepMove, GETNext, SEAGET, ROTAN, CLLP) achieve stronger performance, benefiting from their enhanced temporal modeling capacity. 
Prompt-based methods like LLMMob and ZS-NL demonstrate moderate performance.
LLMMob outperforms ZS-NL by providing more detailed task instructions to guide the LLMs. 
The suffix “s”  indicates a strict version which requires exactly 10 predicted location IDs—if the output count is incorrect,  even if the correct location is predicted, it does not count toward the accuracy.
 This constraint reflects a practical challenge we observed: despite explicit prompt instructions, LLMs frequently produce variable-length outputs, making structured control difficult in prompt-only settings.
This strict enforcement leads to substantial performance drops for both LLMMob and ZS-NL, underscoring the limitations of relying solely on prompting for output structure enforcement.
 Our proposed NextLocLLM outperforms baseline models on most metrics across all datasets, demonstrating strong predictive capabilities.
 Its coordinate regression approach enables better spatial relationship modeling, and LLM-enhanced POI embeddings provide semantic understanding of location functionality, leading to consistent improvements across different datasets.

\subsection{Zero-shot Next Location Prediction Performance}
\label{sec:zero shot}

Table~\ref{tab:zero shot}
presents the performance of different models in zero-shot scenarios. 
Most baselines rely on location IDs, which are city-specific and non-transferable across cities. 
As a result, they cannot be  applied to zero-shot settings.
Prompt-based methods such as LLMMob and ZS-NL, although they also use location IDs, they leverage the pre-trained knowledge and reasoning ability of LLMs to achieve  zero-shot prediction.
Therefore, we compare NextLocLLM with LLMMob and ZS-NL, as these are the  methods capable of zero-shot transfer. 
NextLocLLM outperforms these transferable baselines across all metrics. 
When trained on the dense Xi’an dataset, it achieves the strongest generalization.
Even when trained on the sparse Singapore dataset, NextLocLLM still exceeds the performance of prompt-based methods.
We attribute this strong zero-shot generalization ability to the collection of several architectural design choices.
First,  coordinate-based modeling eliminates the reliance on city-specific location ID systems, allowing the model to generalize  across  cities.
Second, the introduction of prompt prefixes offers high-level task-specific instructions, which enhance the model’s comprehension of prediction objective and trajectory data.
Third, the LLM-enhanced POI embeddings incorporate transferable semantics by encoding natural-language descriptions of POI categories, enabling NextLocLLM to understand functional similarities between locations across cities.
Finally, the partially frozen LLM backbone retains reasoning ability and world knowledge from pretraining, which are essential for understanding trajectory semantics and generalizing to new cities.
Together, these components form a unified and semantically grounded architecture that enables robust performance in unseen cities.

\begin{table}
  \caption{Ablation Study for NextLocLLM Modules on Xi'an}
  \label{tab:abation 1}
  \centering
  \begin{tabular}{|c|c|c|c|c|}
    \toprule
    Prompt Prefix   &  POI & Hit@1 & Hit@5 & Hit@10 \\
    \hline
    $\times$    & $\times$ &  25.81\% & 83.20\% & 97.54\% \\
    $\times$  & \checkmark & 45.79\% & 94.76\% &98.24\% \\
    \checkmark & $\times$  & 39.57\% & 90.81\% & 98.75\% \\
    \hline
    \checkmark  & \checkmark & \textbf{58.14\%} & \textbf{97.14\%} & \textbf{99.36\%} \\
  
    \bottomrule
  \end{tabular}
\end{table}

\subsection{Ablation Study}
\label{sec:ablation}

To validate the effectiveness of different components in NextLocLLM, we conducted a series of ablation experiments in this section. 

\subsubsection{Ablation Study for  Prompt Prefix and POI Semantics}

We first evaluate the impact of  prompt prefix and LLM-enhanced POI embeddings. 
To evaluate the latter, we replace the LLM-enhanced POI embeddings with a simple numerical encoding: each location’s normalized POI category counts is directly mapped through a linear transformation to produce embeddings $E_{POI}\in \mathbb{R}^{d_{poi}}$. Under this setting, Equations (1) and (2) are modified as follows:

\begin{align}
    E_{all_h}&=E_{XY_h}||E_{T_h}||E_{D_h}||E_{Dur_h}||E_{POI_h} \in \mathbb{R}^{M\times (d_{xy}+d_{t}+d_d+d_{dur}+d_{poi})},\\
    E_{all_c}&=E_{XY_c}||E_{T_c}||E_{D_c}||E_{Dur_c}||E_{POI_c} \in \mathbb{R}^{N\times (d_{xy}+d_{t}+d_d+d_{dur}+d_{poi})}.
\label{eq:1}
\end{align}
In this case, since the LLM-enhanced POI embedding is removed, the final trajectory embeddings  $E_{con_h}$ and $E_{con_c}$, are simply  their content embeddings $E_{con_h}$ and $E_{con_c}$.

As shown in Table~\ref{tab:abation 1}, incorporating prompt prefixes consistently improves performance across all metrics, highlighting their role in providing contextual guidance for the LLM.
Furthermore, models equipped with LLM-enhanced POI embeddings outperform those using simple numerical mappings, demonstrating the importance of leveraging natural-language descriptions to better represent location semantics.
Combining both modules yields the highest overall performance, suggesting they play essential roles in supporting semantically grounded and context-aware prediction.

\begin{table}
  \caption{Zero-shot Next Location Prediction Result on Japan}
  \label{tab:zero shot}
  \centering
  \begin{tabular}{|l|ccc|}
    \toprule
     & Hit@1 & Hit@5 & Hit@10 \\ 
     \hline
      LLMMob & 17.63\% & 28.55\% & 37.26\% \\
    LLMMob(s) & 12.26\% & 21.87\% & 31.19\%  \\
    ZS-NL & 13.07\% & 22.31\% & 26.15\% \\
    ZS-NL(s) & 11.32\% & 19.15\% & 23.57\% \\
    NextlocLLM(Singapore->) & 17.85\% & 29.55\% & 39.86\%  \\
    NextlocLLM(Xi'an->) & 18.42\% & 30.26\% & 42.41\% \\
    \bottomrule
  \end{tabular}
\end{table}

\begin{table}
  \caption{Ablation Study for Historical and Current Trajectory on Xi'an}
  \label{tab:abation 3}
  \centering
  \begin{tabular}{|c|c|c|c|c|}
    \toprule
    Historical & Current  &  Hit@1 & Hit@5 & Hit@10 \\
    \hline
     \checkmark  & $\times$ &  18.75\% & 60.62\%  & 82.10\% \\
     $\times$ & \checkmark & 40.23\% & 91.94\% & 98.66\%\\
    \hline
    \checkmark & \checkmark & \textbf{58.14\%} & \textbf{97.14\%} & \textbf{99.36\%} \\

    \bottomrule
  \end{tabular}
\end{table}

\subsubsection{Ablation Study for Historical and Current Trajectory}
To assess the respective contributions of historical and current trajectories, we perform an ablation study by removing each component independently.
As shown in Table~\ref{tab:abation 3}, excluding either trajectory segment leads to a significant drop in performance.
In particular, the model achieves much lower performance when only historical data is used, suggesting that recent movement patterns (current trajectory) are especially important for identifying short-term user intent.
Conversely, using only current records also underperforms the full model, highlighting the complementary role of long-term mobility history in modeling user preferences.
These results confirm that both historical and current trajectories are essential for accurate next-location prediction.

\begin{table}
  \caption{Ablation Study of LLM backbone(fully-supervised scenario)}
  \label{tab:llm usage f}
  \centering
  \begin{tabular}{|l|c|c|c|}
    \toprule
    Fully-supervised(Xi'an) & Hit@1 & Hit @5 & HIt @10 \\ 
    \hline
    NextLocLLM & 58.14\% & 97.14\% & 99.36\% \\
    \hline
    LLM->Transformer & 45.62\% & 83.25\% & 88.78\% \\
    \bottomrule
  \end{tabular}
\end{table}

\begin{table}
  \caption{Ablation Study of LLM backbone (Zero-shot scenario)}
  \label{tab:llm usage z}
  \centering
  \begin{tabular}{|l|c|c|c|}
    \toprule
    Xi'an->Sinapore & Hit@1 & Hit @5 & HIt @10 \\ 
    \hline
    NextLocLLM & 7.256\% & 28.97\% & 32.19\% \\
    \hline
    LLM->Transformer & 3.105\% & 11.45\% & 22.71\% \\
    \bottomrule
  \end{tabular}
\end{table}

\subsubsection{Ablation Study for LLM Backbone}
To understand the contribution of the LLM backbone in NextLocLLM, we conduct an ablation study by replacing it with a randomly initialized Transformer of same architecture and parameter size. 
Despite having similar model capacity, the Transformer-based variant exhibits significantly worse performance in both supervised and zero-shot settings.
We think that this performance degradation stems from two main factors.
First, the pre-trained LLM encodes rich semantic priors and reasoning capabilities accumulated from large-scale textual corpora. 
These capabilities are crucial for understanding trajectory semantics, particularly in scenarios with limited training data or cross-city transfer, where statistical patterns alone are insufficient. Without this foundation, the randomly initialized Transformer struggles to interpret the temporal, spatial, and semantic cues embedded in trajectories.
Second, the LLM-enhanced POI embeddings are derived from the same pre-trained LLM, ensuring a shared semantic space between input features and the prediction backbone. Replacing the LLM disrupts this alignment—since the Transformer lacks the linguistic grounding required to interpret the POI descriptions—thus weakening the model’s ability to leverage textual metadata.

Notably, although the ablated model retains a similar number of parameters, it requires substantially more training epochs to converge and still fails to match the performance of the full model. This highlights that the gains of NextLocLLM are not due to increased model capacity alone, but arise from structurally integrating pre-trained language understanding into  representation and prediction process.
Our design ensures that POI semantics, trajectory context, and prediction logic all interact within a unified semantic space. This alignment enables the model to interpret high-level user intent and generalize across cities—capabilities that purely statistical models or Transformers with same parameter size fail to match.

\begin{table*}[h!]
\centering
\caption{Performance comparison between coordinate and location inputs.}
\label{table:spatial coord}
\begin{tabular}{@{}lccccccc@{}}
\toprule
\textbf{Method}                & \multicolumn{3}{c}{\textbf{Coordinate}} & \multicolumn{3}{c}{\textbf{Location}} \\ 
\cmidrule(lr){2-4} \cmidrule(lr){5-7}
                               & \textbf{Hit@1} & \textbf{Hit@5} & \textbf{Hit@10} & \textbf{Hit@1} & \textbf{Hit@5} & \textbf{Hit@10} \\ 
\midrule
Xi'an (fully supervised)       & 58.14\%        & 97.14\%        & 99.36\%         & 54.57\%        & 85.11\%        & 87.88\%         \\ 
Singapore (fully supervised)   & 7.823\%        & 30.64\%        & 36.15\%         & 5.261\%        & 21.42\%        & 25.54\%         \\ 
Japan & 19.36\% & 31.82\% & 46.06\% & 16.45\% & 25.46\% & 33.41\% \\
\midrule
Singapore->Japan & 17.85\% & 29.55\% & 39.86\% & 0\% & 0\% & 0.02\% \\
Xi'an->Japan & 18.42\% & 30.26\% & 42.41\% & 0\% &0.01\% &0.03\% \\
Japan->Singapore & 6.734\% & 25.89\% & 31.79\% & 0\% & 0\% & 0.01\% \\
Xi'an->Singapore & 7.256\% & 28.97\% & 32.19\% & 0\% & 0\% & 0.01\% \\
\bottomrule
\end{tabular}
\end{table*}

\subsection{Comparative Analysis of Spatial Coordinates and Location IDs}
Table~\ref{table:spatial coord}  compares coordinate-based and location ID-based configuration under both supervised and zero-shot settings.
In the coordinate-based configuration, NextLocLLM takes normalized spatial coordinates as input and directly predicts the coordinates of the next location, which are then mapped to the top-k location IDs using the post-prediction retrieval module.
In the location ID configuration, both the input trajectories and prediction targets are represented using discrete location IDs. 

In the supervised scenario, where the model is trained and evaluated on the same city,  coordinate-based models yield  better results, demonstrating their advantage in capturing spatial continuity and geographic relationships.
In the zero-shot scenario, the gap widens dramatically.
Since location IDs are city-specific,  the ID-based models fail to generalize and produce near-zero accuracy.
In contrast, coordinate-based models maintain solid performance across all transfer directions, thanks to their geometry-aware coordinate-based representation that transfers naturally across cities.
These results confirm that coordinate setting in NextLocLLM not only improves spatial reasoning in single-city settings, but also overcomes the inherent transferability limitations of location ID systems in cross-city generalization.

\subsection{Case study}
\label{sec:case}

\begin{figure}[h]
\centering
\includegraphics[scale=0.13]{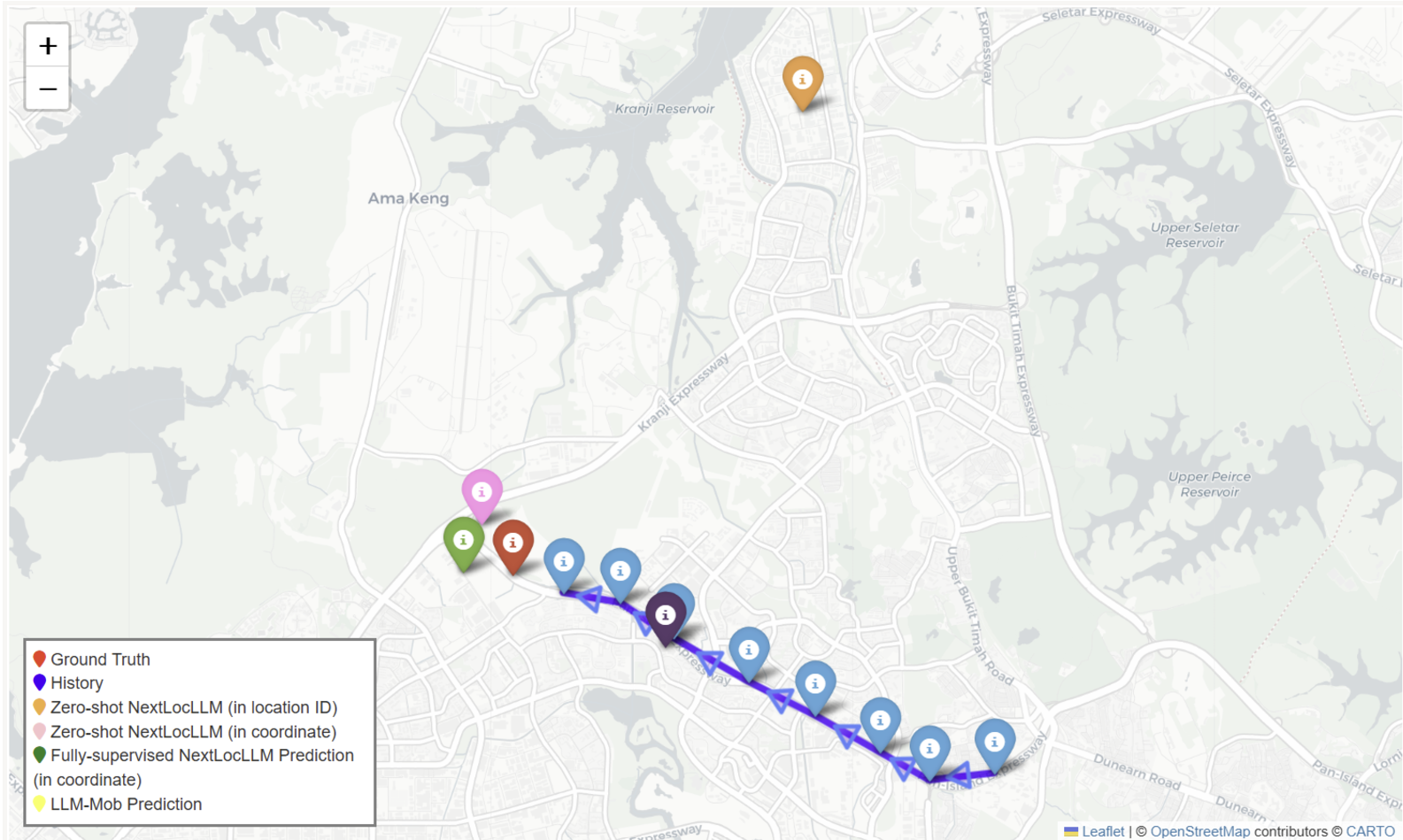}

\caption{Case study on Singapore.  Note that LLM-Mob prediction overlaps exactly with a history visiting record.}
\label{fig:case}
\end{figure}

Fig.~\ref{fig:case} presents a case study evaluating the zero-shot capabilities of NextLocLLM. The model is trained on the Xi’an dataset and tested on a user trajectory from the Singapore dataset. 
For comparison, we also include the fully-supervised setting, where the model is trained and tested on Singapore, along with predictions from the prompt-based LLMMob.
We select Singapore for this case study due to its wide spatial coverage and distinct POI classification system, which differs substantially from Xi’an. 
 This setup mirrors real-world deployment scenarios, where models must generalize across cities with divergent location definitions and semantic taxonomies.

To provide an intuitive comparison, we visualize a single user trajectory with dense sampling (average interval: 21 minutes). The following configurations are included in the figure:
\begin{itemize}
    \item Zero-shot NextLocLLM (using coordinates)
    \item Zero-shot NextLocLLM (using location IDs)
    \item Fully-supervised Next
    LocLLM (using coordinates)
    \item  LLM-Mob prediction
\end{itemize}

\begin{table*}
\caption{Influence of Grid Resolution (Xi'an dataset)}
\label{table:grid}
\begin{tabular}
{@{}lcccccccccc@{}}
\toprule
\textbf{Grid Resolution}                & \multicolumn{3}{c}{\textbf{C-MHSA}} & \multicolumn{3}{c}{\textbf{ROTAN}}& \multicolumn{3}{c}{\textbf{NextLocLLM}} \\ 
\cmidrule(lr){2-4} \cmidrule(lr){5-7} \cmidrule(lr){8-10}
                               & \textbf{Hit@1} & \textbf{Hit@5} & \textbf{Hit@10} & \textbf{Hit@1} & \textbf{Hit@5} & \textbf{Hit@10} 
                                & \textbf{Hit@1} & \textbf{Hit@5} & \textbf{Hit@10}\\ 
\midrule
500m*500m       & 50.32\%        & 92.43\%        & 95.38\%         & 52.36\% & 93.77\% & 96.25\% & 58.14\%        & 97.14\%        & 99.36\%         \\ 
200m*200m   & 40.63\%        & 82.26\%        & 86.77\%  &   44.26\%  &  86.67\% & 88.78\%  & 50.99\%        & 88.42\%        & 94.77\%         \\ 
50m*50m & OOM & OOM  & OOM & OOM & OOM  & OOM & 36.87\% & 62.95\% & 72.47\%  \\
\bottomrule
\end{tabular}
\end{table*}

The experimental results clearly highlight the differences in performance across these methods:
Zero-shot NextLocLLM (using coordinates) produces predictions closely aligned with the ground truth, demonstrating strong generalization to unseen cities. Despite being trained on a different region, the model accurately captures spatial continuity and movement intent.
In contrast, Zero-shot NextLocLLM (using location IDs)  performs significantly worse, deviating from the actual trajectory. This highlights the limitation of using city-specific location IDs, which lack transferability due to inconsistent location ID sets across cities.spaces.
LLM-Mob prediction consistently limits its forecasting to locations  mentioned in the prompts, revealing that models purely based on prompts and location IDs struggle to comprehend spatial information or capture trajectory sequence patterns. 
This limitation significantly reduces the applicability of LLM-Mob for tasks involving unknown locations, whether in zero-shot or fully-supervised scenarios.
Fully-supervised NextLocLLM (using coordinates) serves as an upper bound. Its predictions are only marginally better than the zero-shot version, confirming that NextLocLLM achieves near-supervised performance even without access to target-city training data.

\subsection{Computational Efficiency Analysis}
\begin{table}
  \caption{Training time comparison}
  \label{tab:training time}
  \centering
  \begin{tabular}{|l|c|}
    \toprule
    Model & Training Time (s/iter) \\ 
    \hline
    GRU & 0.131 \\
    \hline
    DeepMove & 0.155 \\
    \hline
    c-MHSA & 0.199 \\
    \hline
    NextLocLLM & 0.935 \\
    \bottomrule
  \end{tabular}
\end{table}

\begin{table}
  \caption{Inference time comparison}
  \label{tab:infer time}
  \centering
  \begin{tabular}{|l|c|}
    \toprule
    Model & Inference Time (s) \\ 
    \hline
    NextLocLLM & 229.2 \\
    \hline
    LLM-Mob & 95232 \\
    \bottomrule
  \end{tabular}
\end{table}

We conduct a detailed analysis of training and inference efficiency for transferable models. 
The results are summarized in Table~\ref{tab:training time} and Table~\ref{tab:infer time}.
During training, NextLocLLM requires 0.935 seconds per iteration, which is higher than GRU (0.131s), DeepMove (0.155s), and c-MHSA (0.199s).
However, the overall computational cost remains acceptable due to the model’s efficient training design.
Most parameters in the LLM backbone are frozen, and only a small set of lightweight modules are updated. 
This significantly reduces the number of trainable parameters and accelerates convergence—on the Xi’an dataset, for example, the model typically converges within 15 epochs.
 As a result, the total training time is manageable, and the substantial performance improvement makes the computational overhead acceptable in practice.

For inference, we compare models that are capable of zero-shot generalization, including NextLocLLM, ZS-NL, and LLMMob. Among them, NextLocLLM demonstrates a clear efficiency advantage, completing inference on the Xi’an dataset in 229.2 seconds, compared to 95,232 seconds for LLMMob. Notably, ZS-NL exhibits a similar inference cost to LLMMob due to their shared reliance on prompt engineering and sequential autoregressive decoding. 
The substantial speedup of NextLocLLM is attributed to two main factors: first, it supports batch inference, allowing simultaneous prediction of multiple trajectories; second, the semantic representations of POI categories are precomputed during data loading and reused throughout inference, eliminating redundant computation and reducing overhead.
While NextLocLLM achieves reasonable computational efficiency, there is still room for improvement. In future work, we plan to explore model compression techniques such as quantization, KV-cache or pruning to reduce runtime costs, as well as knowledge distillation to develop a more lightweight variant suitable for real-time applications.

\subsection{Influence of grid resolution}
\label{sec:grid}
To evaluate how spatial granularity affects prediction accuracy, we conduct experiments on the Xi’an dataset under three grid resolutions: 500m × 500m, 200m × 200m, and 50m × 50m. 
We compare the performance of NextLocLLM with two representative baselines—C-MHSA and ROTAN.
As shown in Table~\ref{table:grid}, all models experience performance degradation as grid resolution increases (as grid size decreases). 
When the grid size reaches 50m × 50m, C-MHSA and ROTAN  run out of memory (OOM) due to the substantial increase in the number of candidate classes. 
This highlights a key limitation of classification-based frameworks: finer spatial resolution dramatically enlarges the label space and resource requirements.
In contrast, NextLocLLM remains computationally feasible under all resolutions. 
Since it directly regresses spatial coordinates rather than classifying over discrete IDs, it avoids the sharp increase in computational cost caused by label space expansion.
Although performance also drops at finer resolutions (e.g., from 58.14\% to 36.87\% Hit@1 when moving from 500m to 50m), NextLocLLM still consistently outperforms both C-MHSA and ROTAN across all settings.
These results validate the robustness of coordinate-based prediction under increasing spatial precision requirements and further support the use of regression-based architectures for fine-grained location modeling. 
However, we also observe that under extremely fine grids, NextLocLLM's performance decline remains notable, suggesting future directions  to further improve precision.

\section{Conclusion}
In this paper, we present NextLocLLM, a novel framework that reformulates the next-location prediction task as a coordinate regression problem, moving away from traditional discrete-location ID classification. 
By directly predicting continuous geographic coordinates, NextLocLLM  maintains spatial continuity and facilitates robust generalization to unseen locations across different cities.
NextLocLLM leverages a unified large language model (LLM) backbone that simultaneously encodes location semantics from natural-language descriptions of POI categories and performs coordinate-level location prediction. 
Specifically, we construct LLM-enhanced POI embeddings to capture functional semantics, and integrate them with spatiotemporal trajectory embeddings within the same LLM-based representation space for coordinate prediction.
This unified representation space ensures that location semantics, trajectory spatial temporal contexts, and prediction logic all interact within one space, enabling LLM to interpret high-level user intent and generalize across semantic similar locations in different cities.
Experimental evaluations across multiple urban datasets demonstrated that NextLocLLM significantly outperforms existing baseline methods in both supervised and zero-shot prediction scenarios. Ablation studies further highlighted the contributions of each core component, confirming the effectiveness of semantic integration, coordinate regression, and LLM backbone.
Despite its strong overall performance, NextLocLLM still faces several limitations.
First, its accuracy drops considerably in fine-grained spatial prediction tasks, where precise localization within dense urban areas remains challenging.
Second, the model’s reliance on high-quality POI textual descriptions may limit its performance in regions with incomplete or noisy semantic information, reducing its ability to capture nuanced location semantics.
Future work will focus on improving regression precision at finer spatial scales, developing strategies to better handle incomplete or noisy POI semantics (e.g., leveraging external knowledge bases or multimodal signals), and further expanding the model’s capability to operate in high-resolution geographic environments.
In summary, NextLocLLM represents a substantial advancement in next-location prediction, offering an effective and generalizable solution with significant potential for diverse real-world applications.

\newpage
\section{GenAI Usage Disclosure}
The proposed model, NextLocLLM, incorporates large language models as part of its methodological framework.
During the preparation of this manuscript, generative AI tools such as ChatGPT were used solely for light editing purposes, including grammar correction, sentence restructuring, and improving the fluency of author-written text. 
No content was generated by AI tools. 
All ideas, methods, experimental analyses, and written content were created entirely by the authors.

\bibliographystyle{unsrt}
\bibliography{sample_base}

\appendix

\end{document}